\documentclass[letterpaper]{article}
\usepackage{algorithmic}
\usepackage{algorithm}
\usepackage{amsmath,amssymb}
\usepackage{amsthm,graphicx,enumerate,verbatim}

\newtheorem{definition}{Definition}
\newtheorem{spec}{Specification}

\begin{document}
\title{Modeling \& Verification of a Multi-Agent Argumentation System using NuSMV}
\author{Supriya D'Souza, Abhishek Rao,~Amit Sharma and Sanjay Singh\thanks {Sanjay Singh is with the Department of Information and Communication Technology, Manipal Institute of Technology, Manipal University, Manipal-576104, INDIA, E-mail: sanjay.singh@manipal.edu}}

\maketitle
%-----------------------------------------------------------------------------------------------------------------------
\begin{abstract} 
Autonomous intelligent agent research is a domain situated at the forefront of artificial intelligence. Interest-based negotiation (IBN) is a form of negotiation in which agents exchange information about their underlying goals, with a view to improve the likelihood and quality of a offer. In this paper we model and verify a multi-agent argumentation scenario of resource sharing mechanism to enable resource sharing in a distributed system. We use IBN in our model wherein agents express their interests to the others in the society to gain certain resources.
\end{abstract}
%\begin{keywords}
%Agent Computing, Argumentation System, Model-Checking, NuSMV
%\end{keywords}
%-----------------------------------------------------------------------------------------------------------------------
\section{Introduction}
Negotiation is a form of interaction in which a group of agents, with conflicting interests, try to come to a mutually acceptable agreement on the distribution of scarce resources. Argumentation-Based Negotiation (ABN) approaches, enable agents to exchange  information (i.e. arguments) during negotiation \cite{ri03}. This paper is concerned with a particular style of argument-based negotiation, namely Interest-Based Negotiation (IBN) \cite{rsd03}, a form of ABN in which agents explore and discuss their underlying interests. Information about other agents' goals may be used in a variety of ways, such as discovering and exploiting common goals.
\par 
Most existing literature supports the claim that ABN is  useful by presenting specific examples that show how ABN can lead to agreement where a more basic exchange of proposals cannot (e.g. the mirror/picture example in \cite{psj98}). The focus is usually on underlying semantics of arguments and argument acceptability. However, no formal analysis exists of how agent preferences, and the range of possible negotiation outcomes,change as a result of exchanging arguments. In this paper, we model and verify a resource sharing mechanism using which agents in a digital ecosystem collaborate.

%-----------------------------------------------------------------------------------------------------------------------	
\section{Preliminaries}
Our negotiation framework consists of a set of two agents \textbf{A} and a finite set of resources  \textbf{R}, which are indivisible. An allocation of resources is a partitioning of \textbf{R} among agents in  \textbf{A} \cite{emst06}.

\begin{definition}[Allocation]
 An allocation of resources  \textbf{R} to 
a set of agents  \textbf{A} is a function $\Lambda : \textbf{A} \rightarrow 2^{\textbf{R}}$ such that $\Lambda (i)\cap
\Lambda(j) = \Phi $ for $i \neq j$ and $\cup_{i \in \textbf{A}} \Lambda(i) = \textbf{R} $.
\end{definition}

Agents may have different preferences over sets of resources,
defined in the form of utility functions.
\begin{definition}[Payment]
 A payment is a function $ p : \textbf{A} \rightarrow \mathbb{R}$
such that  $\sum_{i \in \textbf{A}}p(i) = 0$.
\end{definition}
Note that the definition ensures that the total amount of money is constant. If $p(i) > 0$, the agent pays the amount $p(i)$, while $p(i) < 0$ means the agent receives the amount $-p(i).$ We can now define the notion of 'deal' formally.
\begin{definition}[Deal]
Let $\Lambda$ be the current resource allocation. A deal with money is a tuple $ \Delta = (\Lambda,\Lambda^{'}, p)$ where $\Lambda^{'}$ 
is the suggested allocation, $\Lambda^{'} \neq \Lambda$, and p is a payment.
\end{definition}

%-----------------------------------------------------------------------------------------------------------------------
\section{Methodology}
An offer (or proposal) is a deal presented by one agent which, if accepted by the other agents, would result in a new allocation of resources. In this paper, we will restrict our analysis to two agents. The bargaining protocol initiated by agent $A_i$ with agent $A_j$ is shown in Fig.1.
\par
Bargaining can be seen as a search through possible allocations of resources. In the brute force method, agents would have to exchange every possible offer before a deal is reached or disagreement is acknowledged. The number of possible allocations of resources to agents is $|\textbf{A}|*2^{|\textbf{R}|}$, which is exponential in the number of resources. The number of possible offers is even larger, since agents would have to consider not only every possible allocation of resources, but also every possible payment. Various computational frameworks for bargaining have been proposed in order to enable agents to reach deals quickly \cite{ai07}.

\begin{figure}[H]
	\centering
		\includegraphics[scale=0.5]{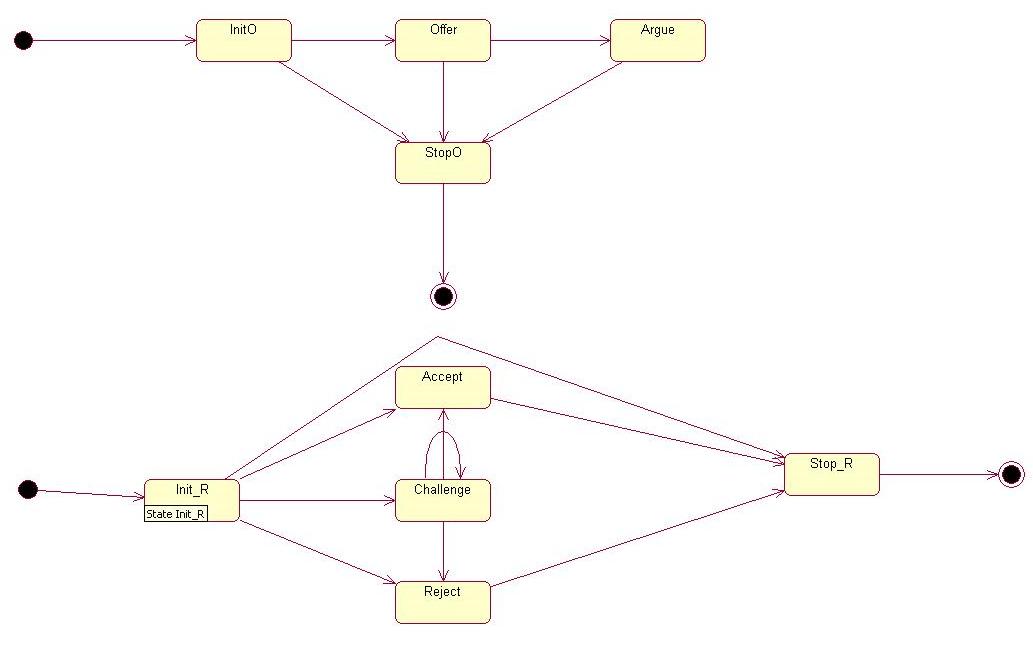}
		\caption{State Chart Diagrams}
	\label{Fig:SD}
\end{figure}

Fig.1 shows state chart diagram which is used to describe the behavior of systems. In Fig.1 above portion show states of offering agent and below part show reacting agent .After initialize both agent offering agent offer some resource to reacting agent, reacting agent either accept,reject or challenge. If reacting agent challenge, so offering agent argue on challenge (A challenge is a continue process till the offering agent does not meet the requirement of reacting agent).Stop state show the termination of communication. \\
%--------------------------------------------------------------------------------------------------------------------
\textbf{Bargaining Protocol(BP):}\\
Agents start with resource allocation $\Lambda^{0}$ at time $t = 0$
At each time $t > 0$:
\begin{enumerate}
\item Propose($A_i$, $\delta^{t}$): Agent $A_i$ proposes to $A_j$ deal $\delta^{t}$ = $(\Lambda^{0},\Lambda^{'},p^{t})$
which has not been proposed before;
\item Agent $A_j$ either:
\begin{enumerate}[(i)]
\item accept(j,$\delta^{t}$): accepts, and negotiation terminates with allocation
$\Lambda^{t}$ and payment $p^{t}$; or
\item reject($A_j$, $\delta^{t}$): rejects, and negotiation terminates with allocation
$\Lambda^{0}$ and no payment; or
\item challenges the argument by going to step 1 at the time step $t + 1$.
\end{enumerate}
\end{enumerate}

%-----------------------------------------------------------------------------------------------------------------------
\section{Model Checking}
Over the years, model checking has evolved greatly into the software domain rather than being confined to hardware such as electronic circuitries. Model checking is one of the most successful approach to verification of any model against formally expressed requirements. It is a technique used for verifying finite state transition system. The specification of system model can be formalized in temporal logic \cite{mm04}, which can be used to verify if a specification holds true in the model. 
\par
Model checking has a number of advantages over traditional approaches which are based on simulation, testing and deductive reasoning. In particular, model checking is an automatic, fast tool to verify the specification against the model. If any specification is false, model checker will produce a counter-example that can be used to trace the source of the error.
\par 
In this paper, we have modeled a resource sharing based argumentation scheme between two agents. In this scenario we have considered a set of resources that are held by the agents. Agents negotiate over the possession of the resources needed by them to achieve their objectives. An Agent wanting a resource makes an initial offer for the resource. The reacting agent or the agent in possession of the resource can either accept, reject or challenge the offer. Based on the move made by the reacting agent the offering agent can either argue or close the dialogue. When an agent accepts a resource from another agent a payment is made to the offering agent.

%----------------------------------------------------------------------------------------------------------------------
\begin{algorithm}[H]
\caption{Offering Agent Behavior}
\label{alg1}
\begin{algorithmic}[1]
\STATE Offering\_Agent()
\STATE ostate = inito
\IF {Offering Agent wants to make an offer }
\STATE ostate = offer
\ENDIF
\IF {Reacting Agent has reached state Accept or Refuse}
\STATE {ostate=stopo}
\ENDIF
\IF {Reacting Agent has challenged the offer made by Offering Agent}
\STATE {ostate = argue}
\ENDIF
\end{algorithmic}
\end{algorithm}

%-----------------------------------------------------------

\begin{algorithm}[H]
\caption{Reacting Agent Behavior}
\label{alg2}
\begin{algorithmic}[1]
\STATE Reacting\_Agent()
\STATE rstate = initr
\IF {Offering Agent has made an offer \& Reacting Agent wants to accept it}
\STATE rstate = accept
\ENDIF
\IF {Offering Agent has made an offer \& Reacting Agent wants to reject it}
\STATE rstate = reject
\ENDIF
\IF {Reacting Agent wants to challenge Offering Agent's offer}
\STATE rstate = challenge
\ENDIF
\IF {Reacting Agent wants to challenge Offering Agent's argument}
\STATE rstate = challenge
\ENDIF
\IF {Offering Agent has gone to stopo state}
\STATE rstate = stopr
\ENDIF
\end{algorithmic}
\end{algorithm}

We have developed two algorithms to demonstrate the behavior of two agents. In algorithm \ref{alg1}, the offering agent makes an offer for a resource. After an offer is made based on the move made by the reacting agent, the offering agent can either argue or stop the dialogue. In algorithm \ref{alg2}, when an offer is made for a resource, the reacting agent can either accept, refuse or challenge the offer.

%-------------------------------------------------------------------------
\section{Verification Results and Discussion}
Properties of the Multi-Agent Argumentation System are specified and evaluated in NuSMV \cite{racgemm}. The system is modeled and fed to the NuSMV tool \cite{ccg02}. We then construct CTL formula, which are in effect, negation of the properties of the system. Each formula is verified by the NuSMV model checker and a counter trace is provided to illustrate that the negated formula are false. We provide the trace after each specification.

\begin{table*}[bpht!]
	\centering
	\caption{Specifications for the Resource Sharing Mechanism}
	\label{SPECs}
	\begin{small}
		\begin{tabular}{|c|c|c|}
			\hline
			\textbf{Sl.No.} & \textbf{Specification} & \textbf{Satisfiability}\\
			\hline
			1 & AG(oagent=offer $\rightarrow$ AF !(ragent=accept|ragent=refuse|ragent=challenge)) & False(Counter-example)\\
			\hline
			2 & AG(ragent=accept $\rightarrow$ AG !(resource[want]=0)) & False(Counter-example)\\
			\hline
			3 & AG(complete $\rightarrow$ AF !(typeChal=0 \& typeArg=0)) & False(Counter-example)\\
			\hline
				\end{tabular}
				\end{small}
	\end{table*}

\begin{spec}
The specification tells that when an offering agent makes an offer, it will neither be accepted, refused nor challenged by the reacting agent. This is FALSE since the reacting agent has to do one of the three options it has. And hence NuSMV generates a counter-example. The Trace shown indicates that reacting agent challenges the offer made by the offering agent.\\
\textit{AG(oagent=offer $\rightarrow$ AF !(ragent=accept|ragent=refuse|ragent=challenge))}.
\end{spec}

\begin{figure}[bpht!]
	\centering
		\includegraphics[scale=0.65]{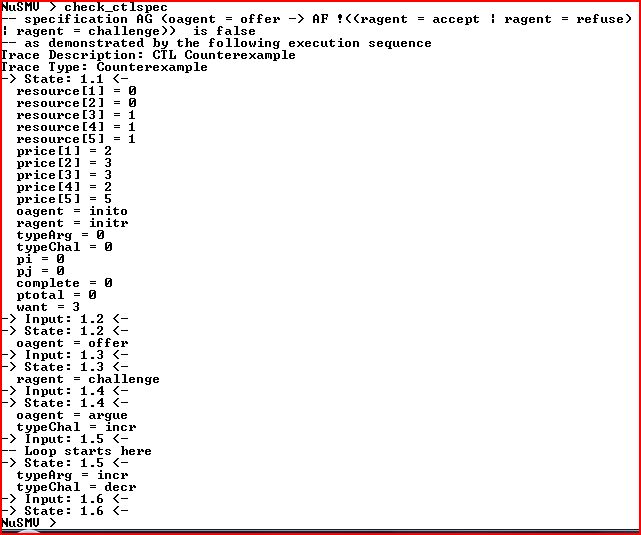}
	\caption{NuSMV Implementation of Specification 1}
	\label{fig:spec1}
\end{figure}

\begin{spec}
The specification tells that, if a reacting agent j reaches a decision to accept the offer, then the resource does not move to the offering agent i. This is FALSE since the resource has to migrate and hence NuSMV generates a counter-example. The Trace indicates that when the offering agent makes an offer for a resource indicated by the variable 'want',when the reacting agent accepts the offer the resource migrates to offering agent and hence its value in not zero.\\
\textit{AG(ragent=accept $\rightarrow$ AG !(resource[want]=0))}.
\end{spec}

\begin{figure}[bpht!]
	\centering
		\includegraphics[scale=0.55]{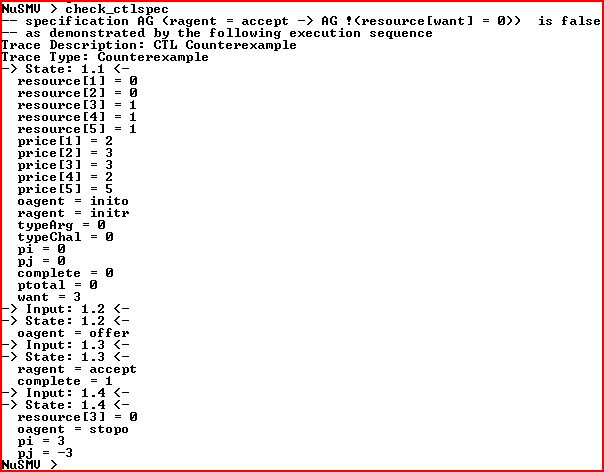}
	\caption{NuSMV Implementation of Specification 2}
	\label{fig:spec2}
\end{figure}

\begin{spec}
The specification tells that,that, More challenges and arguments are made, once a decision has been reached.This is FALSE since no more challenges and arguments are made and hence NuSMV generates a counter-example.The Trace indicates that once the state complete is reached both the offering and reacting agent reach their stop states and hence no more challenges are made.\\
\texttt{AG(complete $\rightarrow$ AF !(typeChal=0 $\land$ typeArg=0))}.
\end{spec}

\begin{figure}[bpht!]
	\centering
		\includegraphics[scale=0.55]{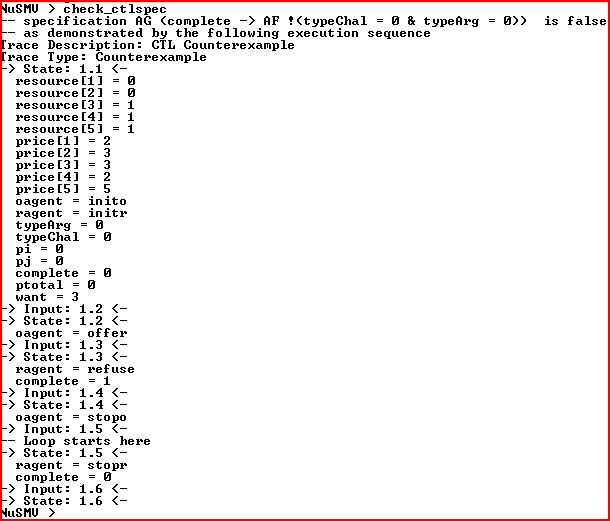}
	\caption{NuSMV Implementation of Specification 3}
	\label{fig:spec3}
\end{figure}
%-----------------------------------------------------------------------------------------------------------------------
\newpage
\section{Conclusion}
In the future, distributed systems will be in the forefront. No distributed system can exist without collaboration. Each distributed system site can have an agent entity to voice its interests. It is not always the case that the interests of all sites will fall in line. This is when argumentation can be useful. In this paper we have demonstrated a simple agent-based argumentation paradigm, where two agents argue on an offer made by one of them, this scenario can be extended for more than two agents. There can be several cycles of challenges and arguments made on the proposal before the agents reach a feasible conclusion. We have modeled the situation and verified it using the NuSMV tool, and the results have been demonstrated. 

\bibliographystyle{IEEEtran}
\bibliography{ref}
\end{document}